\documentclass{article}

\PassOptionsToPackage{numbers, compress}{natbib}

\usepackage[preprint]{neurips_2024}

\usepackage[utf8]{inputenc} 
\usepackage[T1]{fontenc}    
\usepackage{hyperref}       
\usepackage{url}            
\usepackage{booktabs}       
\usepackage{amsfonts}       
\usepackage{nicefrac}       
\usepackage{microtype}      
\usepackage{xcolor}         
\usepackage{graphicx}       
\usepackage{algorithm}
\usepackage{algpseudocode}
\usepackage{multirow}
\usepackage{amsmath}
\usepackage{enumitem}
\usepackage{listings}
\usepackage[figure]{hypcap}

\definecolor{dkgreen}{rgb}{0,0.6,0}
\definecolor{gray}{rgb}{0.5,0.5,0.5}
\definecolor{mauve}{rgb}{0.58,0,0.82}

\title{CORM: Cache Optimization with Recent Message for Large Language Model Inference}

\author{%
  Jincheng Dai \\
  ZheJiang University \\
  \texttt{jincheng\_dai@zju.edu.cn} \\
  \And
  Zhuowei Huang \\
  ZheJiang University \\
  \texttt{zw\_huang@zju.edu.cn} \\
  \And
  Haiyun Jiang\thanks{Corresponding Author} \\
  Tencent AI Lab \\
  \texttt{haiyunjiang@tencent.com} \\
  \AND
  Chen Chen \\
  Tencent AI Lab \\
  \texttt{chenzchen@tencent.com} \\
  \And
  Deng Cai \\
  Tencent AI Lab \\
  \texttt{jcykcai@tencent.com} \\
  \And
  Wei Bi \\
  Tencent AI Lab \\
  \texttt{victoriabi@tencent.com} \\
  \AND
  Shuming Shi \\
  Tencent AI Lab \\
  \texttt{shumingshi@tencent.com} \\
}

\begin{document}

\maketitle

\begin{abstract}
    Large Language Models (LLMs), despite their remarkable performance across a wide range of tasks, necessitate substantial GPU memory and consume significant computational resources. Beyond the memory taken up by model weights, the memory used by the KV cache rises linearly with sequence length, becoming a primary bottleneck for inference. In this paper, we introduce an innovative method for optimizing the KV cache, which considerably minimizes its memory footprint. Upon thorough investigation, we discover that in most Transformer models, (\textit{i}) there is a striking similarity between adjacent tokens' query vectors, and (\textit{ii}) the attention calculation of the current query can rely exclusively on the attention information of a small fraction of preceding queries. Based on these observations, we present CORM, a KV cache eviction policy that dynamically retains essential key-value pairs for inference without the need for model fine-tuning. Our validation shows that CORM reduces the inference memory usage of KV cache by up to 70\% with negligible performance degradation across six tasks in LongBench. Furthermore, we demonstrate that CORM is compatible with GQA for further compression rate.
\end{abstract}

\section{Introduction}
\label{introduction}
Large language models (LLMs) have exhibited remarkable proficiency across a wide range of natural language processing tasks, including question answering, summarization, and multi-turn dialogues \cite{brown2020language, touvron2023llama, thoppilan2022lamda}. However, the considerable deployment cost associated with these models, due to their massive size and quadratic cost of the attention layer, have spurred numerous studies on model compression and memory-efficient attention techniques \cite{dettmers2022llmint8, xia2023sheared, dao2022flashattention, dao2023flashattention}. A critical yet often overlooked aspect is the size of the key-value (KV) cache, which stores prior tokens' key and value states to avoid redundant computation. The KV cache scales linearly with sequence length during generation, resulting in substantial memory overhead. For example, a 7 billion-parameter model with a batch size of 128 and a sequence length of 4096 necessitates a 256GB KV cache. This far exceeds the memory consumed by the model itself, which amounts to a mere 14GB. One intuitive solution is to eliminate less informative KV cache to decrease memory usage. The primary challenge, however, is striking a delicate balance between discarding as much KV cache as possible while preserving optimal model performance.

Despite multi-query attention \cite{shazeer2019fast} and grouped-query attention \cite{ainslie2023gqa} can reduce the size of KV cache by reducing key-value heads, these methods need retraining to restore performance of original model. Recent studies \cite{xiao2023efficient, liu2024scissorhands, zhang2024h2o, oren2024transformers, ren2024efficacy} have explored implementing KV cache using specific eviction policy, that determines which key-value states should be evicted from KV cache. These methods are geared towards condensing the KV cache to a predefined budget size, thereby reducing memory and computational overhead. However, they preserve same quantity of key-value pairs across all attention heads and layers, neglecting that the number of keys playing an important role may vary across different attention heads and layers, as highlighted in \cite{michel2019sixteen}. 

\begin{figure}[!h]
  \centering
  \setlength{\abovecaptionskip}{-0.4cm}
  \includegraphics[width=\linewidth]{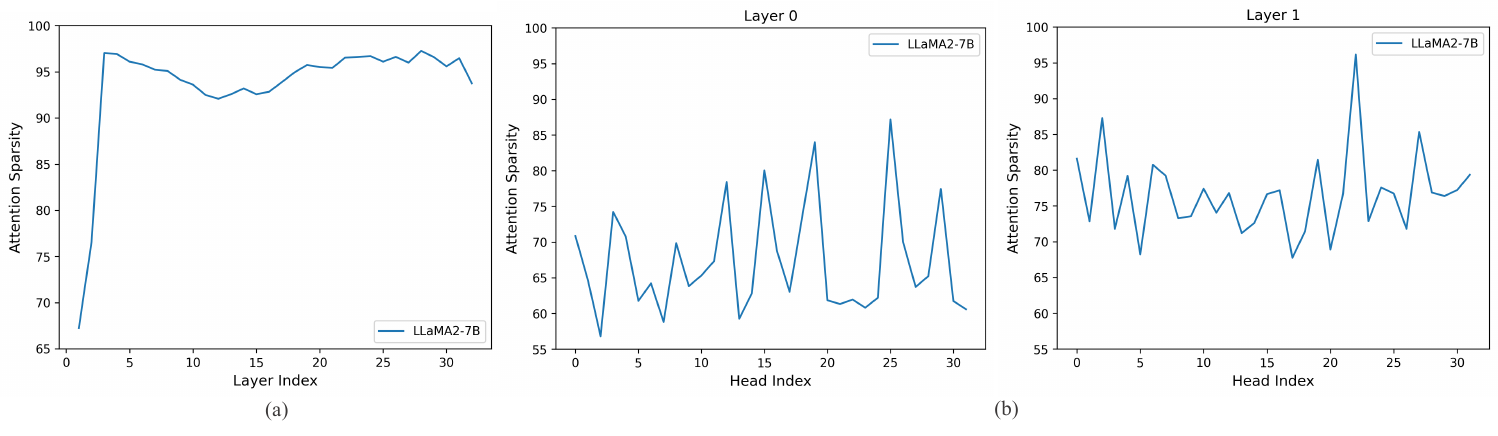}
  \caption{Attention sparsity of LLaMA2-7B. (a) Layer-wise attention sparsity. (b) Head-wise attention sparsity of layer 0 and layer 1.}
  \label{attention_sparsity}
\end{figure}

Intuitively, in scenarios where critical information within KV cache exceeds the predefined budget size, it is reasonable to anticipate a degradation in model performance due to the inevitable eviction of crucial information. Our preliminary investigations corroborate this intuition, revealing distinct degrees of sparsity across different attention layers and heads as depicted in Figure \ref{attention_sparsity}. First, we observe that bottom layers of the model are relatively dense\footnote{Let $t$ denote current decoding step, we count the proportion of keys which attention score larger than average score $\frac{1}{t}$ and denote it as $r$. A higher value of $r$ indicates greater sparsity within the layer and head.}, while the remaining attention layers exhibit significant sparsity. Second, even within the same layer, different heads can exhibit obvious differences in sparsity degrees. These properties suggest that we need to allocate budget for different layers and heads individually, rather than applying the same budget size across the entire model. In addition, we prove that completely similar queries have similar concerns for keys, and observe that recent query vectors are quite similar on Transformer models. Consequently, it is feasible for the current query to leverage recent query attention messages during generation process.

Based on the above insights, we propose \textbf{C}ache \textbf{O}ptimization with \textbf{R}ecent \textbf{M}essage (CORM), a framework that exploits recent query attention messages for KV cache optimization and token generation of LLMs. Specifically,

\begin{itemize}[leftmargin=*, topsep=0pt,parsep=0pt,partopsep=0pt]
\item In Section \ref{observation}, we explore the similarity between query vectors of all tokens within same sequence, revealing that recent query vectors are highly similar, which implies that (\textit{i}) keys that are important for recent queries might be also important for the current query; and (\textit{ii}) removing key-value pairs that appear to be less informative for recent queries can greatly preserve the performance of the model.

\item In Section \ref{method}, we present a simple method which dynamically evicts minor key-value pairs determined by recent query attention messages, and design the generation process algorithm of LLMs with a budget-unrestricted KV cache.
\end{itemize}

We conduct comprehensive experiments on LLaMA2-7B-Chat and Vicuna-7b-v1.5-16k, considering their popularity and extensive usage, to evaluate CORM across 6 tasks from LongBench \cite{bai2023longbench} containing question answering, summarization, code completion, etc. Experiments indicate that even without explicitly setting a budget size, CORM is still able to achieve a high compression rate. Our method achieves better performance compared to StreamingLLM, Scissorhands and H$_2$O with over 70\% KV cache reduction rate and can even come close to fully restoring the performance of the model. Furthermore, we demonstrate that CORM can be effectively integrated with grouped-query attention (GQA) to achieve addictional compression without noticeable performance degradation.

\section{Related Work}
\label{related_work}
\paragraph{Attention} Let $x \in \mathbb{R}^{n \times d}$ denote the input embeddings from a sequence of $n$ feature vectors of dimension $d$. The multi-head self-attention \cite{vaswani2017attention}, as a core module of Transformer model, facilitates contextual information interaction within each head in the following manner:
\begin{equation}
\textit{Attention}(x) = \textit{softmax}(\frac{QK^T}{\sqrt{d_h}})V.
\end{equation}

Q, K, V represent the query, key, and value matrices, which are obtained by linearly mapping $x$ using weight matrices $W_q$, $W_k$, and $W_v \in \mathbb{R}^{d \times d_{h}}$, respectively. $d_h$ is the dimension of each individual head. 

\paragraph{KV Cache Compression} A common paradigm for inference of Transformer models is to retain the key-value pairs of previous tokens for subsequent reuse, to avoid inefficiency of recomputation. Thus, the consumption of KV cache becomes linearly correlated with sequence length, potentially resulting in excessive memory and latency issues when dealing with long input or output.

Many efforts have been made to improve efficiency for LLMs. Multi-query attention (MQA) \cite{shazeer2019fast} and grouped-query attention (GQA) \cite{ainslie2023gqa} are proposed to reduce key-value heads to decrease memory usage and the number of memory swaps. Adaptively Sparse Attention \cite{anagnostidis2024dynamic} learns to drop uninformative tokens from the context at any point across the generation process. Dynamic Memory Compression (DMC) \cite{nawrot2024dynamic} learns to apply different compression rates in different heads and layers with continual training. However, these methods require additional training to keep model performance due to the incapacity of direct conversion.

Another set of efforts is dedicated to addressing the balance between model efficiency and inference cost, achieved without extra training and architectural changes, by evicting the KV cache using different algorithms. StreamingLLM \cite{xiao2023efficient} and LM-Infinite \cite{han2023lm} keep initial token and recent tokens throughout decoding process to align with the training window. InfLLM \cite{xiao2024infllm} stores evicted tokens into additional memory units and employs an efficient mechanism to lookup token-relevant units for attention computation. Scissorhands \cite{liu2024scissorhands} maintains pivotal tokens and recent tokens based on the persistence of importance hypothesis. H$_2$O \cite{zhang2024h2o} utilizes accumulated attention score to maintain heavy hitters and recent tokens for inference. TOVA \cite{oren2024transformers} removes tokens with the lowest current attention score from the fixed cache at each decoding step. RoCo \cite{ren2024efficacy} retains tokens in the fixed cache based on high mean cumulative attention scores and top r standard deviations. SnapKV \cite{li2024snapkv} automatically compresses KV cache by selecting clustered important KV positions for each attention head. Aforementioned methods consistently operate on a fixed cache, ignoring that the number of tokens playing an important role may vary across different attention heads and layers. Though FastGen \cite{ge2023model} applies four compression policies to construct KV cache in an adaptive manner, human-imposed strategies and model behaviors may exhibit certain biases, ultimately impacting model performance.

\section{Observations}
\label{observation}
We first demonstrate \textit{the existence of attention sparsity in LLMs} in Section \ref{Section 3.1}, then discuss the phenomenon that \textit{similar queries have similar attention concerns for keys} in Section \ref{Section 3.2}. In Section \ref{Section 3.3}, we show an intriguing observation that \textit{current query is most similar to recent queries}.

\subsection{Attention sparsity in LLMs}
\label{Section 3.1}

We first explore the sparsity in attention layers of LLMs, which provides an effective guarantee for us to reduce KV cache size. Specifically, we use proportion of important keys to represent attention sparsity. Let $q_{t} \in \mathbb{R}^{1 \times d}$ denote the query state vector at step $t$, $k_{i} \in \mathbb{R}^{1 \times d}$ denote the key state vector at step $i$ $(1 \le i \le t)$, where $d$ is hidden dimension (for the sake of simplicity, we only consider a single head here). The normalized attention score of $q_{t}$ for $k_{i}$ is computed as:

\begin{equation}
    \alpha_{t,i} = \frac{\mathrm{exp}(q_{t} k^{T}_{i} / \sqrt{d})}{\sum_{j=1}^{t}\mathrm{exp}(q_{t} k^{T}_{j} / \sqrt{d})}.
\end{equation}

\paragraph{Definition 3.1 \normalfont{(Important Key)}} We define a key $k_{i}$ is considered important at step $t$, if and only if $\alpha_{t,i} \ge \frac{1}{t}$, otherwise it is considered minor.

We conduct zero-shot inference with LLaMA2-7B model on the test set of PG-19 \cite{rae2019compressive}. We plot the layer-wise and head-wise sparsity within attention blocks, the figures are presented in Figure \ref{attention_sparsity}. It reveals that attention score matrices of the model is really sparse, specifically the bottom layers are relatively dense, while other layers are highly sparse with over 90\% sparsity. This makes it possible to do attention computation on only small part of KV cache during generation. We also give attention sparsity plots of Falcon-7B and Qwen1.5-7B in Appendix \ref{appendix_sparsity}.

\subsection{Similar queries have similar concerns for keys}
\label{Section 3.2}

The previous section reveals the existence of attention sparsity in LLMs, which provides an opportunity to reduce KV cache size while maintaining performance. In this section we provide both theoretical analysis and empirical evidence that \textit{similar queries have similar concerns for keys} for eviction policy design.

Consider the $i$-th and $j$-th query state vectors $q_{i}$ and $q_{j}$ in a sequence of token length $T$ ($i < j \le T$). Their cosine similarity can be computed as:
\begin{equation}
\label{cosine_similarity}
    \textit{cosine\_similarity}(q_{i}, q_{j}) = \frac{q_{i} q_{j}^T}{\left \| q_{i} \right \| \cdot \left \| q_{j} \right \| }.
\end{equation}

Consider all key states $k_{1}$, $k_{2}$, ..., $k_{i-1}$ before $i$-th key. Assume that $\textit{cosine\_similarity}(q_{i}, q_{j})=1$, then $q_{i}=m \cdot q_{j}$ where $m \in \mathbb{R}^+$. The attention weight\footnote{Attention weight is unnormalized attention score.} of $q_{i}$ to the previous $i-1$ keys can be represented as:

\begin{equation}
\label{distribution_invariant}
\textit{attention\_weight} = \frac{1}{\sqrt{d}} \cdot (q_{i}k_{1}^T, q_{i}k_{2}^T, ..., q_{i}k_{i-1}^T) = \frac{m}{\sqrt{d}} \cdot (q_{j}k_{1}^T, q_{j}k_{2}^T, ..., q_{j}k_{i-1}^T).
\end{equation}

Note that $m$ is a positive number that does not affect the relative order of the attention weights. For example, if $q_{i}k_{1}^T > q_{i}k_{2}^T$ for $q_{i}$, there must be $q_{j}k_{1}^T > q_{j}k_{2}^T$ for $q_{j}$. 
This means if a key is important to $q_{i}$, it is also important to $q_{j}$, though the degree of importance may vary due to the softmax function.

\begin{figure}[!h]
  \centering
  \setlength{\abovecaptionskip}{-0.4cm}
  \includegraphics[width=\linewidth]{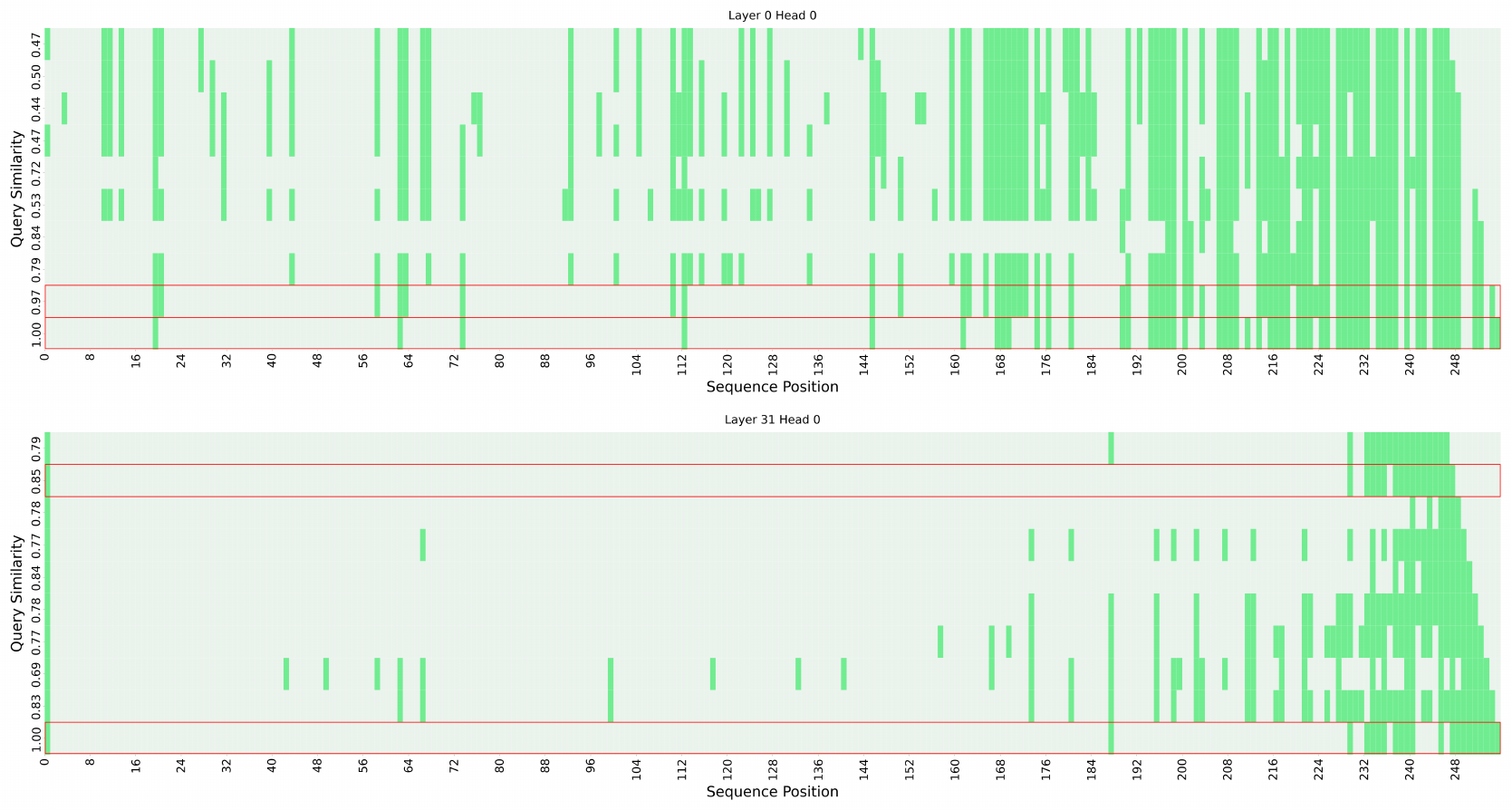}
  \caption{Similar queries have similar concerns for keys. We plot the attention maps from two different layers in a sentence. We discretize the attention score and those important keys are shown in bright green. Each attention map has two red borders, the bottom border shows important keys that current query actually focuses on, while another border shows important keys that the most similar query focuses on.}
  \label{sim_importance}
\end{figure}

Although it's nearly impossible that $\textit{cosine\_similarity}(q_{i}, q_{j})=1$ in real situation, we can make the hypothesis that two similar queries may have similar concerns for keys. 
To validate this hypothesis, we provide two attention maps of a sentence randomly drawn from PG-19 using LLaMA2-7B, as shown in Figure \ref{sim_importance}. Important keys are marked with bright green, more plots are available in Appendix \ref{appendix_token_similarity}. 
We observe that the hypothesis is true, similar queries exhibit similar concerns for important keys. At the same time, important keys only account for a small proportion especially in deeper attention layers, which is consistent with the finding that deeper layers are sparser in previous section.

\subsection{Similarity exploration of query vectors}
\label{Section 3.3}
We have validated \textit{two similar queries have similar concerns for keys} in Section \ref{Section 3.2}, we further
investigate whether at each step we can find a previous query state that is similar enough to current query state in same layer and same head. To check this, we visualize cosine similarity of query vectors within same sequence as shown in Figure \ref{query_relationship}, more plots are available in Appendix \ref{appendix_query_similarity}. We observe an intriguing phenomenon that many images show clear oblique color segmentation, with the top oblique block closest to dark red which means current query is most similar to recent queries.

\begin{figure}[!h]
  \centering
  \setlength{\abovecaptionskip}{-0.4cm}
  \includegraphics[width=\linewidth]{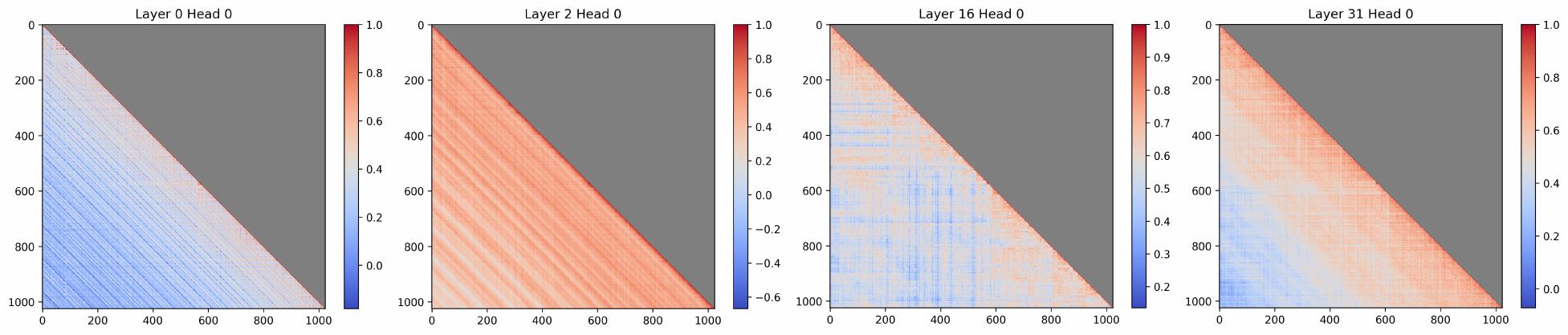}
  \caption{Visualization of query vectors' cosine similarity over randomly sampled sentence with a length of 1024 on LLaMA2-7B. The $i$-th row of the map represents cosine similarity of the $i$-th query to all previous queries. The redder the color, the higher the similarity between two queries. The plot reveals that in most cases current query is most similar to recent queries.}
  \label{query_relationship}
\end{figure}

Through above observations, we see an opportunity to design a KV cache eviction policy based on query similarity that preserves LLMs generation performance. 

\section{Cache Optimization with Recent Message}
\label{method}

In this section, we present CORM, a method reduces KV cache memory based on recent query attention messages without any fine-tuning process. In Section \ref{Section 4.1}, we derive that current query can directly use recent query attention messages during generation. In Section \ref{Section 4.2}, we present CORM eviction policy and describe how it works during generation. 

\subsection{Inference based on recent query attention messages}
\label{Section 4.1}
Consider observations in Section \ref{observation}, intuitively, we can directly store all queries and their attention messages for future reference. At each generation step, use current query to find the most similar one from previous queries, and use its saved attention information to calculate solely on important keys. However, this approach incurs a significant additional cost. First, storing all queries results in a substantial increase in memory overhead. Second, the requirement of performing similarity calculations at each step adds to the computational overhead.

Since in most cases current query is most similar to recent queries as described in Section \ref{Section 3.3}, we can just use recent query attention messages. And from Figure \ref{sim_importance} we can also observe that only a small proportion of keys are considered important by recent queries. Therefore even if we save all the keys that are considered important in previous steps, we can save a lot of memory. 

\subsection{Eviction algorithm with recent message}
\label{Section 4.2}
We have shown recent query attention information is enough for cache optimization in previous section. In the following, we formally define this algorithm and introduce how to integrate it into LLM generation directly.

\paragraph{Definition 4.1 \normalfont{(Long-term Minor Key)}} A key $k_{i}$ is considered as long-term minor key only if it is considered minor in all recent $r$ steps (from $t-r+1$ to $t$).

\paragraph{Approach} CORM will have a recent window of size $w$ to record the information of recent $w$ queries (for each query, we will record the keys it considers important), and will always keep recent $r$ keys unremoved to prevent them from being discarded prematurely due to insufficient observations. During generation, $k_{i}, v_{i}$ will be discarded once $k_{i}$ is regarded as a long-term minor key. For better explanation we present PyTorch code\footnote{For the sake of brevity, the code snippet only demonstrates single-head eviction operation, while in the actual implementation, it will be performed on each head at every layer.} of main algorithm in Algorithm \ref{algorithm1}. Intuitively, when $w$ is larger, more keys and values will be saved, the compression rate will be smaller and performance will be better; Conversely, when $w$ is smaller, fewer keys and values will be saved, the compression rate will be larger and performance will be worse. So there's a tradeoff between performance and compress rate.

\paragraph{Memory Overhead Analysis} In order to reduce memory overhead of KV cache, an extra memory overhead is introduced by recent information cache. We need to store recent query messages which increase memory overhead. However, these overheads are far less than compressed KV cache, one can use a small portion of memory to avoid maintaining full KV cache memory without obvious performance degradation.

\begin{algorithm}
\caption{Single-head KV cache eviction with CORM (unbatched)}
\label{algorithm1}
\begin{lstlisting}
def corm_eviction(keys, values, message, attn_score, w, r, t):
"""
Args:
    keys: previous key states, a tensor with shape [l, d]
    values: previous value states, a tensor with shape [l, d]
    message: attention message, a tensor with shape of [m, l-1]
    attn_score: current step's attention score, a tensor with shape of [1, l]
    w: window size, a scalar
    r: recent size, a scalar
    t: current step, a scalar
Returns:
    updated_keys: updated keys
    updated_values: updated values
    updated_message: updated message
"""
m = message.shape[0]

# update attention message
message = torch.cat([message, torch.zeros(m, 1)], dim=1)  # pad to [m, l] 
cur_message = attn_score >= 1/t
message = torch.cat([message, cur_message], dim=1)[-w:, :]

if message.shape[0] < w:
    return keys, values, message
else:
    # determine the key-value pairs that necessitate discarding
    decision = message.any(dim=0)
    decision[-r:] = True   # always keep recent r tokens unremoved
    indices = torch.nonzero(decision).squeeze()
    
    keys = keys[indices, :]
    values = values[indices, :]    
    return keys, values, message
\end{lstlisting}
\end{algorithm}

\section{Empirical Evaluation}
\label{evaluation}
In this section, we present the results that demonstrate CORM can reduce up to 70\% of the memory footprint of KV Cache without accuracy degradation on LLaMA2-7B-Chat and Vicuna-7b-v1.5-16k. 

\paragraph{Dataset} To broadly validate feasibility of our method on real-world use cases, we choose LongBench \cite{bai2023longbench} as our evaluation benchmark, which contains a wide range of long-text tasks such as question answering \cite{kovcisky2018narrativeqa, dasigi2021dataset, yang2018hotpotqa, ho2020constructing, trivedi2022musique, he2017dureader}, summarization \cite{huang2021efficient, zhong2021qmsum, fabbri2019multi, wu2023vcsum}, few-shot learning \cite{li2002learning, lsht, gliwa2019samsum, joshi2017triviaqa}, synthetic task and code completion \cite{guo2023longcoder, liu2023repobench}. Here we do not consider short text tasks, because even full cache doesn't have any memory bottlenecks.

\paragraph{Models} Since sequence length is the main factor in the continuous growth of KV Cache, we employ LLaMA2-7B-Chat \cite{touvron2023llama} for 4K test and Vicuna-7b-v1.5-16k for 12k test considering their wide usage.

\paragraph{Baselines} Since CORM reduces KV cache without need for training, we consider several similar approaches as our baselines: StreamLLM \cite{xiao2023efficient}, Scissorhands \cite{liu2024scissorhands} and H$_2$O \cite{zhang2024h2o}. In addition, the full KV cache is also considered as strong baseline to measure performance loss of other methods.

\begin{figure}[!h]
  \centering
  \setlength{\abovecaptionskip}{-0.4pt}
  \includegraphics[width=0.85\linewidth]{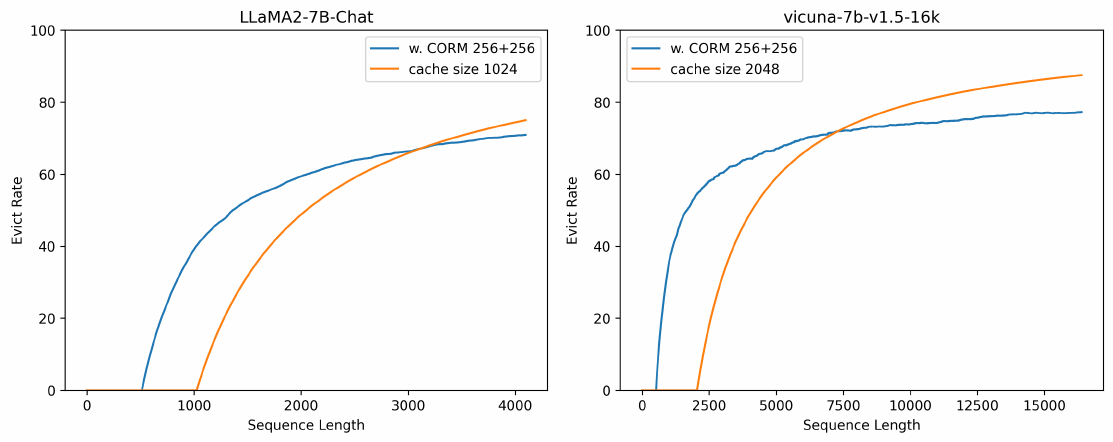}
  \caption{Relationship between compression rate and sequence length averaged by 10 texts randomly sampled from PG19. Plots show that compression rate with CORM "256+256" (w=256, r=256) closely matches a budget of 1024 for LLaMA2-7B-Chat, and a budget of 2048 for Vicuna-7b-v1.5-16k.}
  \label{ratio-len}
\end{figure}

\paragraph{Settings} All baselines can be regarded as fixed budget size KV cache compression, however CORM is a dynamic compression method. Since we find that CORM has similar compression rates for various task texts with the same sequence length. For fair comparison, we plot the relationship between model compression rate and sequence length using texts randomly sampled from PG19 \cite{rae2019compressive} as shown in Figure \ref{ratio-len}. For LLaMA2-7B-Chat we use NVIDIA V100 32GB GPU and for Vicuna-7b-v1.5-16k we use NVIDIA A100 80GB GPU.

\paragraph{Main Results} We evaluate LLaMA2-7B-Chat for 4K length text and Vicuna-7b-v1.5-16k for 12k length text. Results are summarized in Table \ref{llama2_result1} \& \ref{llama2_result2} for LLaMA2-7B-Chat and Table \ref{vicuna_result1} \& \ref{vicuna_result2} for Vicuna-7b-v1.5-16k. The following observations can be drawn: (1) CORM consistently outperforms previous methods at the similar compression rate across a wide range of tasks. (2) Meanwhile, with over 70\% KV cache reduction, CORM achieves comparable performance as the model with full KV cache and even surpass it on some tasks, we speculate it's because there's some noise in full KV cache that affects model output and our method can eliminate this noise to a certain extent by discarding some key-value pairs. We also evaluate the perplexity of different methods on first 5 texts drawn from PG-19 test set, the result is shown in Table \ref{ppl_result}, it also shows CORM is better than other methods.

\begin{table}[!ht]
  \caption{Results (\%) on single-doc QA, multi-doc QA and summarization tasks. "Full" refers to LLaMA2-7B-Chat utilizing full KV Cache, "StreamLLM" is configured with 4+1020, "Scissorhands" is configured with 768+256 where window size=256, "H$_2$O" is configured with 768+256, "CORM" is configured with 256+256 for fair comparison. For the sake of brevity we use ID to denote dataset here, mapping from ID to dataset can be found in Appendix \ref{appendix_task_mapping}.}
  \centering
  \resizebox{\linewidth}{!}{
      \begin{tabular}{l*{12}{c}}
        \toprule
        \multirow{2}{*}{Method}     & \multicolumn{4}{c}{Single-Doc QA}     & \multicolumn{4}{c}{Multi-Doc QA}     & \multicolumn{4}{c}{Summarization}\\
        \cmidrule(r){2-5}\cmidrule(r){6-9}\cmidrule(r){10-13}
        & 1-1 & 1-2 & 1-3 & 1-4 & 2-1 & 2-2 & 2-3 & 2-4 & 3-1 & 3-2 & 3-3 & 3-4 \\
        \midrule
        Full & 19.0 & 22.1 & 36.7 & 11.8 & 27.8 & 31.5 & 8.3 & 6.8 & 26.8 & 20.7 & 26.2 & 0.2 \\
        StreamLLM & 13.2 & 15.4 & 27.2 & 6.5 & 24.2 & 25.4 & 5.3 & 4.4 & 21.6 & 19.8 & 24.4 & 0.1 \\
        Scissorhands & 16.6 & 18.7 & 32.4 & 9.9 & 26.3 & 32.1 & 8.9 & 5.7 & 22.1 & 20.7 & 25.4 & 0.2 \\
        H$_2$O & 17.9 & 19.5 & 34.9 & 11.5 & 27.5 & 29.7 & 7.5 & 7.1 & 24.5  & 21.0 & 25.8 & 0.2 \\
        CORM & 18.9 & 22.2 & 38.6 & 12.0 & 27.6 & 31.6 & 8.4 & 7.1 & 26.4 & 21.0 & 25.8 & 0.2 \\
        \bottomrule
      \end{tabular}
    }
  \label{llama2_result1}
\end{table}

\begin{table}[!ht]
  \caption{Results (\%) on few-shot learning, synthetic, and code tasks. "Overall" is computed by the macro-average over major task categories. This is computed on English (EN) tasks, Chinese (ZH) tasks, and all (All) tasks, code tasks are included in both languages.}
  \centering
  \resizebox{\linewidth}{!}{
      \begin{tabular}{l*{12}{c}}
        \toprule
        \multirow{2}{*}{Method}     & \multicolumn{4}{c}{Few-shot Learning}     & \multicolumn{3}{c}{Synthetic}     & \multicolumn{2}{c}{Code} & \multicolumn{3}{c}{Overall} \\
        \cmidrule(r){2-5}\cmidrule(r){6-8}\cmidrule(r){9-10}\cmidrule(r){11-13}
        & 4-1 & 4-2 & 4-3 & 4-4 & 5-1 & 5-2 & 5-3 & 6-1 & 6-2 & EN & ZN & All \\
        \midrule
        Full & 64.0 & 83.3 & 41.4 & 17.3 & 2.9 & 7.8 & 10.0 & 58.3 & 52.2 & 32.8 & 16.9 & 28.9 \\
        StreamLLM & 61.0 & 82.9 & 39.1 & 14.5 & 1.8 & 4.7 & 6.5 & 57.6 & 50.0 & 29.5 & 14.3 & 25.7 \\
        Scissorhands & 52.5 & 83.6 & 40.7 & 17.0 & 3.1 & 6.5 & 7.7 & 56.8 & 52.1 & 31.0 & 15.8 & 27.2 \\
        H$_2$O & 63.0 & 81.5 & 39.9 & 17.0 & 2.8 & 7.0 & 7.3 & 57.8 & 52.3 & 31.8 & 16.4 & 28.0 \\
        CORM & 64.0 & 83.5 & 41.3 & 17.3 & 2.9 & 9.0 & 9.1 & 58.3 & 52.0 & 32.9 & 16.8 & 28.9 \\
        \bottomrule
      \end{tabular}
    }
  \label{llama2_result2}
\end{table}

\subsection{Impact of window size}
Intuitively, when window size is larger, more key-value pairs will be cached which means more information will be saved, the performance will be better. So we explore the impact of window size on model performance by evaluating on LLaMA2-7B-Chat under different window sizes. We present the results in Appendix \ref{ablation_window_size}, it shows that there is a positive correlation between window size and model performance. However, even if we set the window size to 64, it won't bring much performance degradation, which means we can keep model performance with even higher compression rate.

\begin{table}[!ht]
  \caption{Perplexity comparison of different models and different methods.}
  \centering
      \begin{tabular}{l*{5}{c}}
        \toprule
            Model &  Full  & StreamLLM & Scissorhands & H$_2$O & CORM \\   
        \midrule
        LLaMA2-7B & 9.79 & 10.15 & 10.05 & 10.02 & 9.89 \\
        Vicuna-7b-v1.5-16k & 12.49 & 13.15 & 13.80 & 154.02 & 12.88 \\
        \bottomrule
      \end{tabular}
  \label{ppl_result}
\end{table}

\begin{table}[!ht]
  \caption{Different Positional Encoding 128 + 128 2048.}
  \centering
      \begin{tabular}{l*{5}{c}}
        \toprule
            Model &  Full & CORM & Reduction\\   
        \midrule
        LLaMA1-7B & 10.58 & 10.69 & 74.4\% \\
        Bloom-7B & 17.02 & 17.05 & 4.4\% \\
        OPT-6.7B & 15.83 & 16.39 & 73.4\% \\
        \bottomrule
      \end{tabular}
  \label{pos_embedding}
\end{table}

\subsection{Budget unnecessity}
We primarily focus on the effectiveness of not setting a budget versus setting a fixed budget. Note that since we use same window size and recent size as Scissorhands in the experiment, it can be regarded as a natural ablation experiment. And Table \ref{llama2_result1} \& \ref{llama2_result2} have shown that, at the similar compression rate, CORM is much better than Scissorhands on most tasks, and performance of other tasks is close. This verifies that different transformer layers and heads should be treated differently rather than setting a same fixed budget size or a single compression rate. We attribute this to the varying sparsity across different layers and heads of Transformer models.

\subsection{Integrate into grouped-query attention (GQA)}
GQA is widely adopted in LLMs to reduce memory overhead, we also extend CORM to GQA, for different groups, we discard keys that are considered minor by all recent queries within the group. The results of LLaMA3-8B-Instruct in Table \ref{llama3_result1} \& \ref{llama3_result2} validate the effectiveness of our extension, even in GQA setting, CORM can still achieve 60\% KV cache reduction and keep performance.

\begin{table}[!ht]
  \caption{Results (\%) on single-doc QA, multi-doc QA and summarization tasks. "Full" refers to LLaMA3-8B-Instruct utilizing full KV Cache, "CORM" is configured with 256+256.}
  \centering
  \resizebox{\linewidth}{!}{
      \begin{tabular}{l*{12}{c}}
        \toprule
        \multirow{2}{*}{Method}     & \multicolumn{4}{c}{Single-Doc QA}     & \multicolumn{4}{c}{Multi-Doc QA}     & \multicolumn{4}{c}{Summarization}\\
        \cmidrule(r){2-5}\cmidrule(r){6-9}\cmidrule(r){10-13}
        & 1-1 & 1-2 & 1-3 & 1-4 & 2-1 & 2-2 & 2-3 & 2-4 & 3-1 & 3-2 & 3-3 & 3-4 \\
        \midrule
        Full & 21.0 & 44.1 & 44.9 & 51.8 & 47.2 & 35.8 & 22.9 & 13.0 & 29.1 & 21.6 & 24.5 & 0.2 \\
        CORM & 21.0 & 43.7 & 44.8 & 51.3 & 45.7 & 36.3 & 23.1 & 13.7 & 28.6 & 21.7 & 24.4 & 0.2 \\
        \bottomrule
      \end{tabular}
    }
  \label{llama3_result1}
\end{table}

\begin{table}[!ht]
  \caption{Results (\%) on few-shot learning, synthetic, and code tasks.}
  \centering
  \resizebox{\linewidth}{!}{
      \begin{tabular}{l*{12}{c}}
        \toprule
        \multirow{2}{*}{Method}     & \multicolumn{4}{c}{Few-shot Learning}     & \multicolumn{3}{c}{Synthetic}     & \multicolumn{2}{c}{Code} & \multicolumn{3}{c}{Overall} \\
        \cmidrule(r){2-5}\cmidrule(r){6-8}\cmidrule(r){9-10}\cmidrule(r){11-13}
        & 4-1 & 4-2 & 4-3 & 4-4 & 5-1 & 5-2 & 5-3 & 6-1 & 6-2 & EN & ZN & All \\
        \midrule
        Full & 74.5 & 90.5 & 42.3 & 23.8 & 6.0 & 64.5 & 85.5 & 56.9 & 51.6 & 42.6 & 38.1 & 42.2 \\
        CORM & 74.5 & 90.6 & 41.9 & 23.8 & 6.0 & 64.0 & 82.5 & 57.9 & 52.3 & 42.6 & 37.8 & 42.0 \\
        \bottomrule
      \end{tabular}
    }
  \label{llama3_result2}
\end{table}

\subsection{Generalization}
\label{generalization}
CORM works on the observation that \textit{similar queries have similar concerns for keys}. We further explore whether all Transformer models have this phenomenon. We conduct experiments on Falcon-7B \cite{refinedweb}, Mistral-7B \cite{jiang2023mistral}, LLaMA3-8B, Qwen1.5-7B \cite{qwen}, OPT-6.7B \cite{zhang2022opt} and Bloom-7B \cite{workshop2022bloom}. 
We plot query similarity within same sequence as shown in Figure \ref{phenomenon}, more plots can be found in Appendix \ref{simq_diff_visual}. The results show that each model has such phenomenon except OPT-6.7B and Bloom-7B.
We find that OPT-6.7B use absolute position embedding and Bloom-7B use ALiBi \cite{press2021train} position embedding, while others use rotary position embedding (RoPE) \cite{su2024roformer}. 
We hypothesis that rotary position embedding can allow the model to make adjacent queries more similar during training process. 
We conduct further experiments to compare the perplexity of LLaMA1-7B, OPT-6.7B, and Bloom-7B on the first five texts of PG19 as shown in Table \ref{pos_embedding}, it reveals that there are two situations when CORM is applied to non-RoPE models: (\textit{i}) CORM cannot effectively reduce the space occupied by KV cache, for example the compression rate of Bloom-7B is only 4\% after applying CORM. (\textit{ii}) CORM can effectively reduce the space occupied by KV cache, such as OPT-6.7B, but the performance loss will be higher compared to RoPE models.

\begin{figure}[!ht]
  \centering
  \setlength{\abovecaptionskip}{-0.4cm}
  \includegraphics[width=\linewidth]{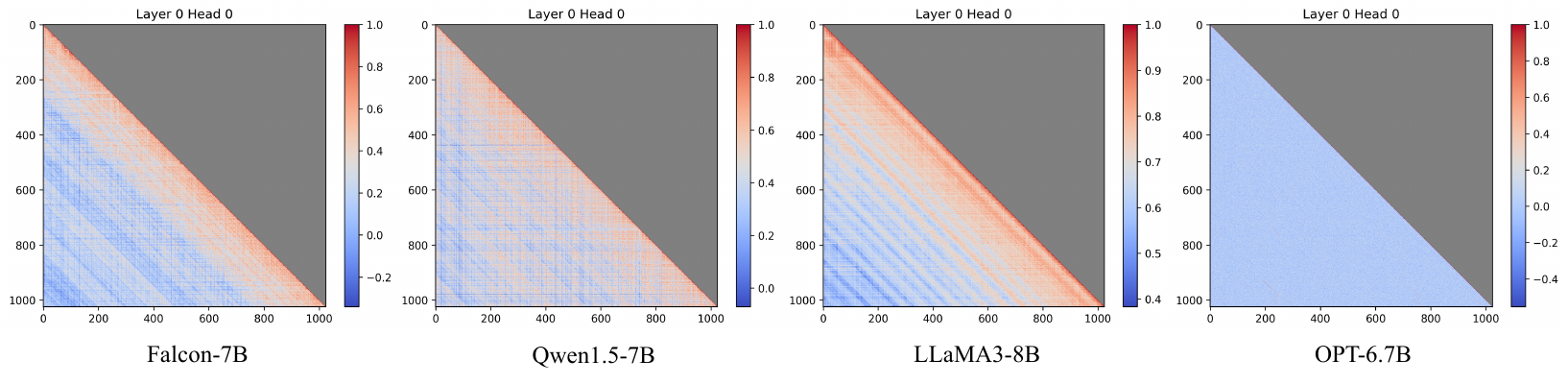}
  \caption{Visualization of query vectors' cosine similarity over randomly sampled sentence with a length of 1024 across Falcon-7B, Qwen1.5-7B, LLaMA3-8B, OPT-6.7B.}
  \label{phenomenon}
\end{figure}

\section{Conclusion}
In this paper, we address a critical memory bottleneck in deployment of LLM, specifically KV cache. Drawing on the observation that similar queries exhibit similar concerns for keys, and recent queries tend to be sufficiently similar, we introduce CORM, an innovative eviction policy for KV cache that does not rely on a predefined budget. It significantly diminishes the memory footprint by utilizing attention messages of recent queries. Through comprehensive evaluations, we demonstrate that CORM can reduce the inference memory usage of KV cache by up to 70\%, while maintaining robust performance across a variety of tasks.

\newpage

\bibliographystyle{unsrtnat}
\bibliography{Ref}

\newpage
\appendix


\section{More Plots}
\label{appendix_plots}

\subsection{Attention sparsity in LLMs}
\label{appendix_sparsity}
\begin{figure}[ht]
  \centering
  \setlength{\abovecaptionskip}{-0.4cm}
  \includegraphics[width=\textwidth]{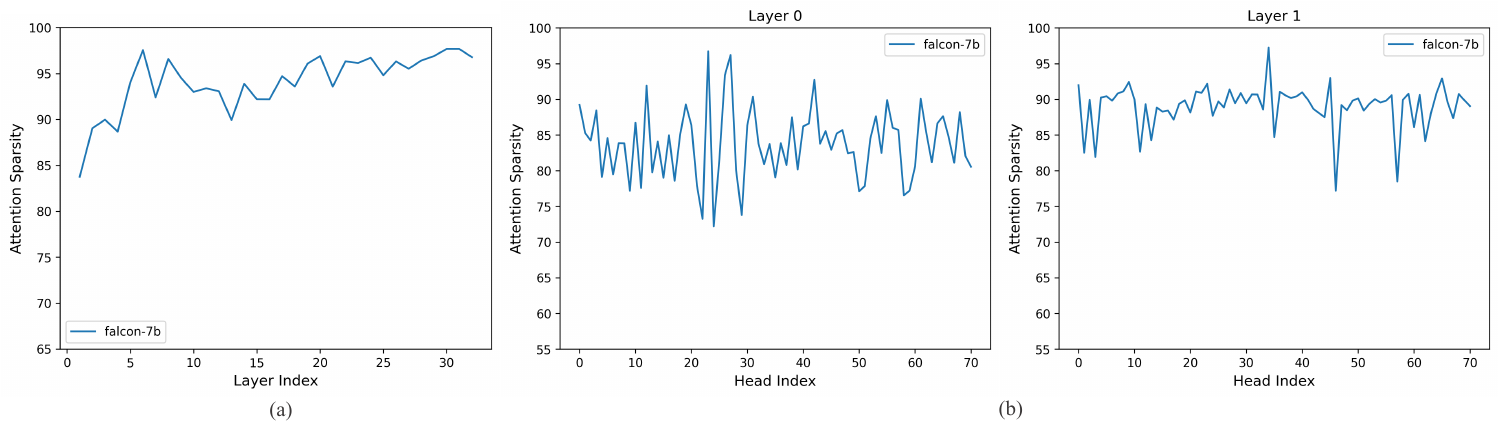}
  \caption{Attention sparsity of Falcon-7B. (a) Layer-wise attention sparsity. (b) Head-wise attention sparsity of layer 0 and layer 1.}
  \label{falcon_sparisty}
\end{figure}

\begin{figure}[ht]
  \centering
  \setlength{\abovecaptionskip}{-0.4cm}
  \includegraphics[width=\textwidth]{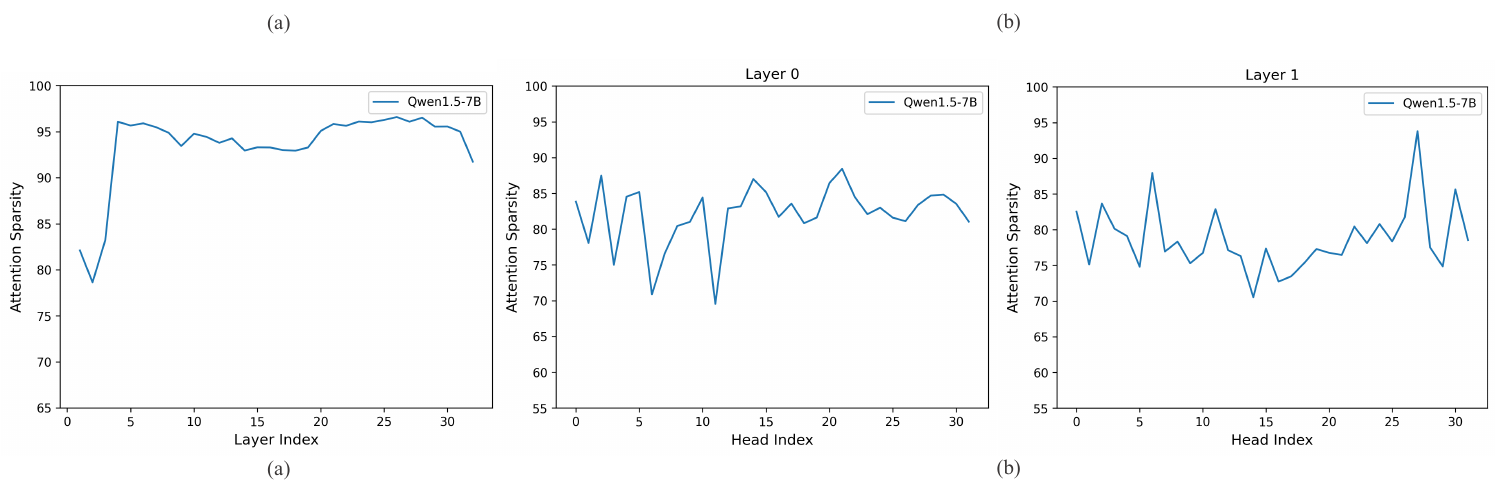}
  \caption{Attention sparsity of Qwen1.5-7B. (a) Layer-wise attention sparsity. (b) Head-wise attention sparsity of layer 0 and layer 1.}
  \label{qwen_sparisty}
\end{figure}

We provide attention sparsity visualizations of Falcon-7B and Qwen1.5-7B, as depicted in Figure \ref{falcon_sparisty} and Figure \ref{qwen_sparisty}, respectively. They also demonstrate that the attention score matrices exhibit sparsity, with varying degrees of sparsity observed across different attention layers and heads.

\subsection{Similar queries have similar concerns for keys}
\label{appendix_token_similarity}
\begin{figure}[ht]
  \centering
  \setlength{\abovecaptionskip}{-0.4cm}
  \includegraphics[width=\textwidth]{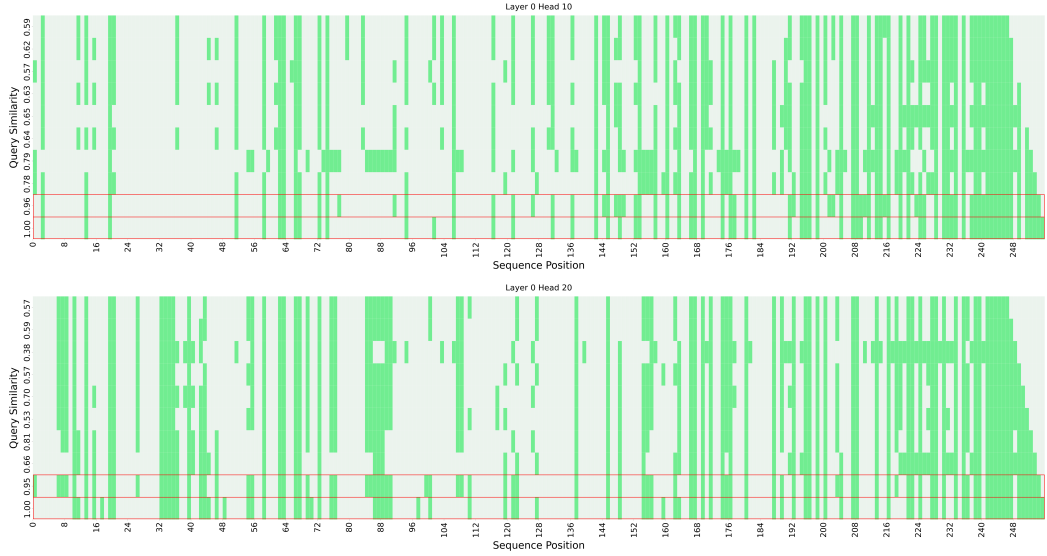}
  \caption{Attention Map at Layer 0, Head 10, 20}
  \label{layer0_different_head}
\end{figure}

\begin{figure}[!ht]
  \centering
  \setlength{\abovecaptionskip}{-0.4cm}
  \includegraphics[width=\textwidth]{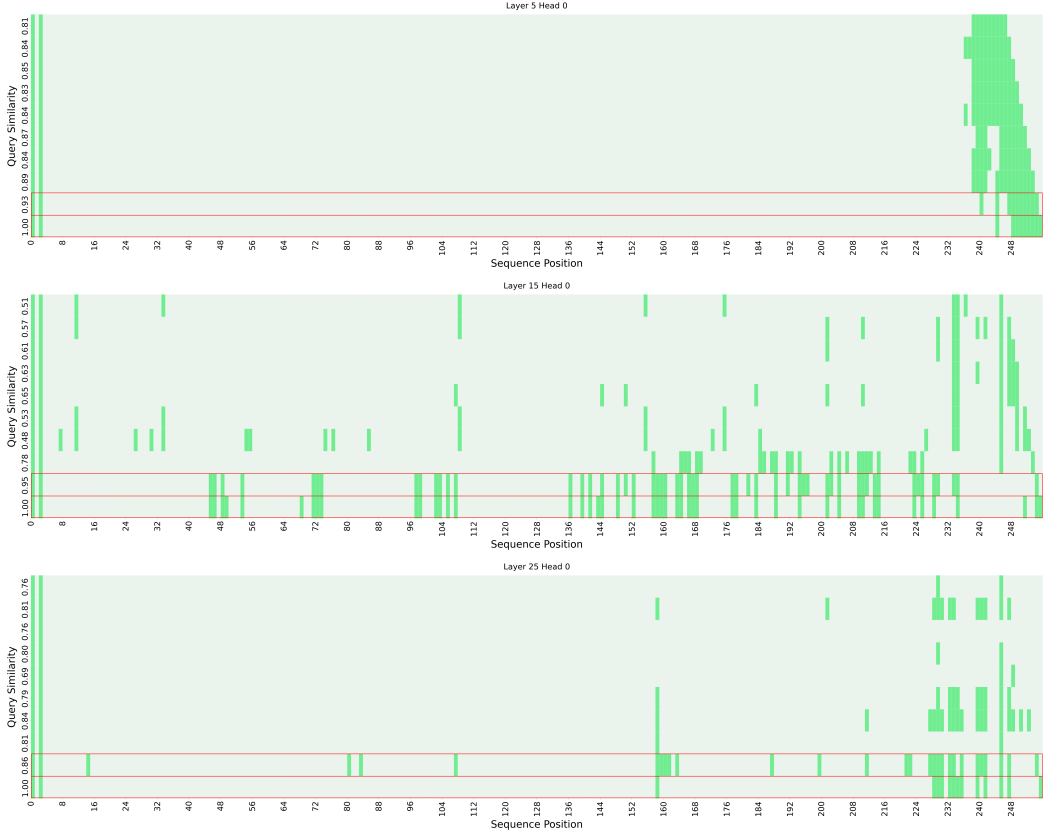}
  \caption{Attention Map at Layer 5, 15, 25, Head 0}
  \label{different_layer_head0}
\end{figure}

We provide the attention maps similar to Figure \ref{sim_importance} but from different heads on the same text in Figure \ref{layer0_different_head}, Figure \ref{layer31_different_head}. Plots from a different layer on the same text are shown in Figure \ref{different_layer_head0}.

\begin{figure}[ht]
  \centering
  \setlength{\abovecaptionskip}{-0.4cm}
  \includegraphics[width=\textwidth]{figures/appendix\_figure2.pdf}
  \caption{Attention Map at Layer 31, Head 10, 20}
  \label{layer31_different_head}
\end{figure}

\subsection{Similarity between query vectors}
\label{appendix_query_similarity}
\begin{figure}[!ht]
  \centering
  \setlength{\abovecaptionskip}{-0.4cm}
  \includegraphics[width=\textwidth]{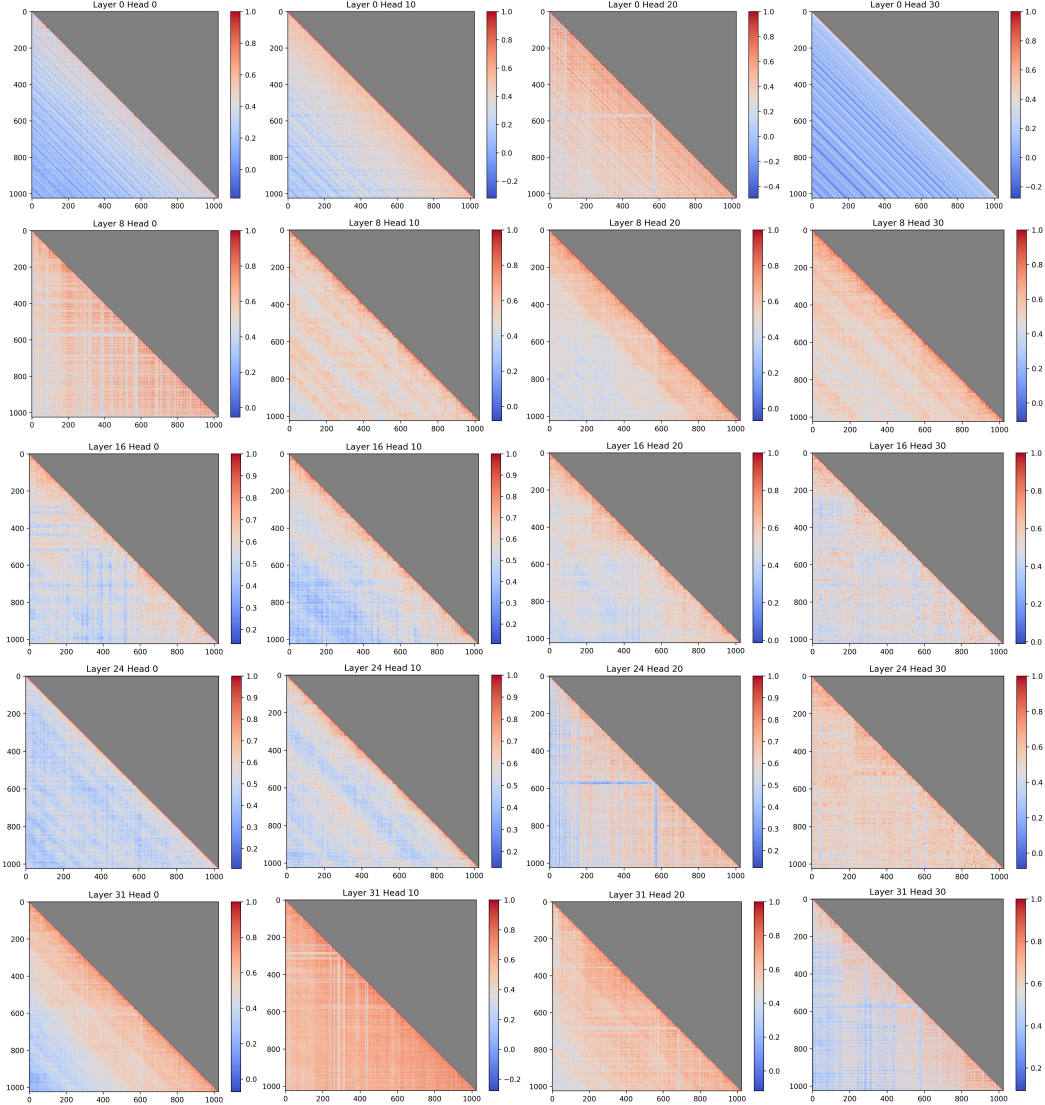}
  \caption{Visualization of query vectors’ cosine similarity across different layers and heads}
  \label{query_similarity}
\end{figure}

We provide the query vectors' cosine similarity visualizations similar to Figure \ref{query_relationship} but from different layers and heads on the same text in Figure \ref{query_similarity}.

\section{Task Mapping}
\label{appendix_task_mapping}
\begin{figure}[!ht]
  \centering
  \setlength{\abovecaptionskip}{-0.4cm}
  \includegraphics[width=\textwidth]{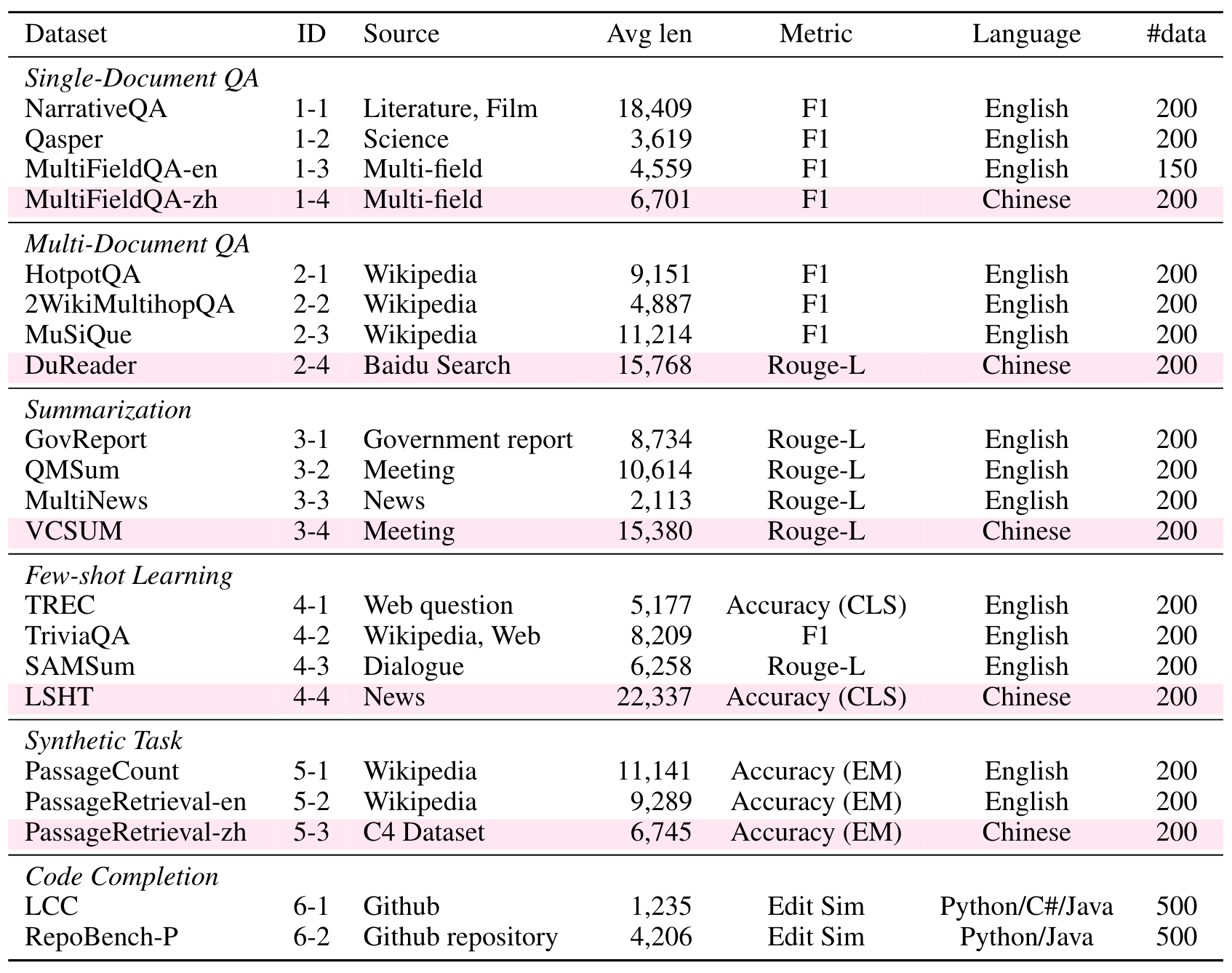}
  \caption{An overview of the dataset statistics in LongBench}
  \label{longbench_dataset}
\end{figure}

An overview of the dataset statistics and mapping from ID to dataset in LongBench is shown in Figure \ref{longbench_dataset}.

\section{Additional Experiment}
\subsection{Vicuna results}
We evaluate Vicuna-7b-v1.5-16k for 12k length text, results are summarized in Table \ref{vicuna_result1} \& \ref{vicuna_result2}. "Full" refers to Vicuna-7b-v1.5-16k utilizing full KV Cache, "StreamLLM" is configured with 4+2044, "Scissorhands" is configured with 1792+256 where window size=256, "H$_2$O" is configured with 1792+256, "CORM" is configured with 256+256 for fair comparison. 

\begin{table}[!ht]
  \caption{Results (\%) on single-doc QA, multi-doc QA and summarization tasks.}
  \centering
  \resizebox{\linewidth}{!}{
      \begin{tabular}{l*{12}{c}}
        \toprule
        \multirow{2}{*}{Method}     & \multicolumn{4}{c}{Single-Doc QA}     & \multicolumn{4}{c}{Multi-Doc QA}     & \multicolumn{4}{c}{Summarization}\\
        \cmidrule(r){2-5}\cmidrule(r){6-9}\cmidrule(r){10-13}
        & 1-1 & 1-2 & 1-3 & 1-4 & 2-1 & 2-2 & 2-3 & 2-4 & 3-1 & 3-2 & 3-3 & 3-4 \\
        \midrule
        Full & 16.6 & 26.5 & 38.4 & 41.8 & 23.3 & 21.4 & 8.6 & 23.3 & 29.0 & 22.7 & 27.5 & 14.5 \\
        StreamLLM  & 12.1 & 20.6 & 20.0 & 29.2 & 19.1 & 18.4 & 4.4 & 16.1 & 13.2 & 18.0 & 25.9 & 13.1 \\
        Scissorhands & 14.1 & 22.0 & 28.8 & 25.8 & 19.4 & 21.7 & 5.5 & 13.3 & 20.7 & 19.5 & 26.8 & 8.0 \\
        H$_2$O  & 10.9 & 23.2 & 29.7 & 31.6 & 16.8 & 16.8 & 6.4 & 19.3 & 25.6 & 21.2 & 27.2 & 13.7 \\
        CORM & 16.5 & 25.5 & 36.4 & 41.3 & 25.2 & 22.0 & 7.7 & 22.7 & 27.4 & 23.1 & 26.9 & 13.2 \\
        \bottomrule
      \end{tabular}
    }
  \label{vicuna_result1}
\end{table}

\begin{table}[!ht]
  \caption{Results (\%) on few-shot learning, synthetic, and code tasks.}
  \centering
  \resizebox{\linewidth}{!}{
      \begin{tabular}{l*{12}{c}}
        \toprule
        \multirow{2}{*}{Method}     & \multicolumn{4}{c}{Few-shot Learning}     & \multicolumn{3}{c}{Synthetic}     & \multicolumn{2}{c}{Code} & \multicolumn{3}{c}{Overall} \\
        \cmidrule(r){2-5}\cmidrule(r){6-8}\cmidrule(r){9-10}\cmidrule(r){11-13}
        & 4-1 & 4-2 & 4-3 & 4-4 & 5-1 & 5-2 & 5-3 & 6-1 & 6-2 & EN & ZN & All \\
        \midrule
        Full & 69.5 & 86.5 & 41.0 & 27.3 & 4.0 & 5.0 & 4.5 & 49.2 & 43.2 & 31.3 & 26.3 & 30.0 \\
        StreamLLM & 62.0 & 84.5 & 29.6 & 19.8 & 3.0 & 5.5 & 6.0 & 46.5 & 39.3 & 26.1 & 21.2 & 24.9\\
        Scissorhands  & 54.0 & 71.8 & 38.6 & 19.5 & 5.0 & 2.5 & 2.2 & 48.1 & 32.6 & 26.4 & 18.2 & 24.3\\
        H$_2$O & 67.0 & 79.9 & 28.6 & 17.2 & 4.7 & 4.0 & 2.6 & 50.0 & 36.4 & 27.6 & 21.3 & 26.0\\
        CORM & 69.0 & 83.7 & 41.3 & 25.3 & 4.5 & 5.5 & 5.0 & 48.9 & 43.4 & 31.0 & 25.6 & 29.7\\
        \bottomrule
      \end{tabular}
    }
  \label{vicuna_result2}
\end{table}

\subsection{CORM with different window sizes}
\label{ablation_window_size}
We evaluate CORM on LLaMA2-7B-Chat with different window sizes, results are summarized in Table \ref{ablation_llama1} \& \ref{ablation_llama2}. 

\begin{table}[!ht]
  \caption{Results (\%) on single-doc QA, multi-doc QA and summarization tasks.}
  \centering
  \resizebox{\linewidth}{!}{
      \begin{tabular}{l*{12}{c}}
        \toprule
        \multirow{2}{*}{Method}     & \multicolumn{4}{c}{Single-Doc QA}     & \multicolumn{4}{c}{Multi-Doc QA}     & \multicolumn{4}{c}{Summarization}\\
        \cmidrule(r){2-5}\cmidrule(r){6-9}\cmidrule(r){10-13}
        & 1-1 & 1-2 & 1-3 & 1-4 & 2-1 & 2-2 & 2-3 & 2-4 & 3-1 & 3-2 & 3-3 & 3-4 \\
        \midrule
        Full & 19.0 & 22.1 & 36.7 & 11.8 & 27.8 & 31.5 & 8.3 & 6.8 & 26.8 & 20.7 & 26.2 & 0.2 \\
        CORM 256+256 & 18.9 & 22.2 & 38.6 & 12.0 & 27.6 & 31.6 & 8.4 & 7.1 & 26.4 & 21.0 & 25.8 & 0.2 \\
        CORM 128+128 & 18.8 & 21.5 & 38.4 & 11.8 & 27.0 & 31.9 & 8.3 & 6.7 & 24.7 & 20.9 & 25.5 & 0.2 \\
        CORM 64+64 & 19.3 & 20.9 & 37.8 & 10.8 & 27.3 & 31.6 & 8.6 & 6.6 & 22.5 & 21.2 & 24.3 & 0.2 \\
        CORM 32+32 & 19.1 & 20.1 & 37.3 & 10.8 & 27.7 & 31.7 & 8.5 & 5.6 & 21.0 & 21.2 & 22.2 & 0.2 \\
        CORM 16+16 & 18.8 & 18.7 & 35.4 & 9.6 & 28.1 & 31.0 & 8.1 & 5.6 & 20.2 & 20.8 & 20.1 & 0.1 \\
        \bottomrule
      \end{tabular}
    }
  \label{ablation_llama1}
\end{table}

\begin{table}[!ht]
  \caption{Results (\%) on few-shot learning, synthetic, and code tasks.}
  \centering
  \resizebox{\linewidth}{!}{
      \begin{tabular}{l*{12}{c}}
        \toprule
        \multirow{2}{*}{Method}     & \multicolumn{4}{c}{Few-shot Learning}     & \multicolumn{3}{c}{Synthetic}     & \multicolumn{2}{c}{Code} & \multicolumn{3}{c}{Overall} \\
        \cmidrule(r){2-5}\cmidrule(r){6-8}\cmidrule(r){9-10}\cmidrule(r){11-13}
        & 4-1 & 4-2 & 4-3 & 4-4 & 5-1 & 5-2 & 5-3 & 6-1 & 6-2 & EN & ZN & All \\
        \midrule
        Full & 64.0 & 83.3 & 41.4 & 17.3 & 2.9 & 7.8 & 10.0 & 58.3 & 52.2 & 32.8 & 16.9 & 28.9 \\
        CORM 256+256 & 64.0 & 83.5 & 41.3 & 17.3 & 2.9 & 9.0 & 9.1 & 58.3 & 52.0 & 32.9 & 16.8 & 28.9 \\
        CORM 128+128 & 64.0 & 83.0 & 41.2 & 17.3 & 3.4 & 8.5 & 8.5 & 57.2 & 52.7 & 32.7 & 16.6 & 28.7\\
        CORM 64+64 & 64.0 & 83.0 & 40.7 & 17.8 & 3.0 & 8.0 & 8.0 & 58.4 & 52.1 & 32.4 & 16.4 & 28.5\\
        CORM 32+32 & 64.0 & 83.4 & 40.5 & 16.8 & 3.6 & 7.0 & 7.7 & 57.6 & 51.3 & 32.0 & 15.9 & 28.0\\
        CORM 16+16 & 63.5 & 83.4 & 40.1 & 18.0 & 2.9 & 5.5 & 6.6 & 54.7 & 50.6 & 31.0 & 15.4 & 27.2 \\
        \bottomrule
      \end{tabular}
    }
  \label{ablation_llama2}
\end{table}

\subsection{Query similarity visualization of different models}
\label{simq_diff_visual}
\begin{figure}[!ht]
  \centering
  \setlength{\abovecaptionskip}{-0.4cm}
  \includegraphics[width=\textwidth]{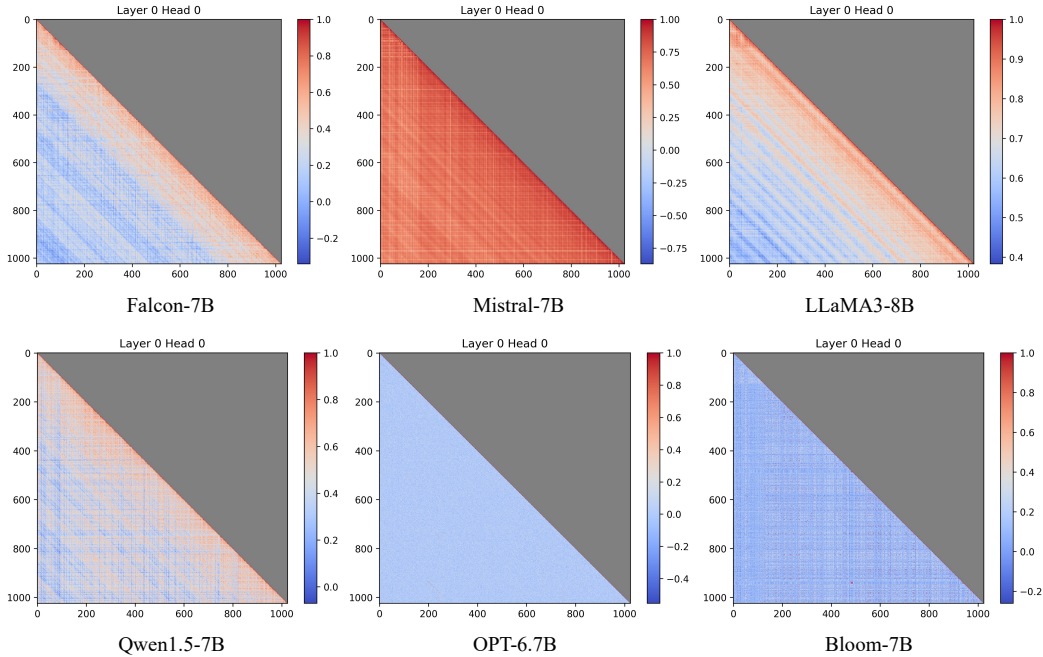}
  \caption{Visualization of query vectors’ cosine similarity over randomly sampled sentence with a length of 1024 across Falcon-7B, Mistral-7B, LLaMA3-8B, Qwen1.5-7B, OPT-6.7B, Bloom-7B}
  \label{simq_diff_fig}
\end{figure}

We plot query similarity within same sequence across Falcon-7B, Mistral-7B, LLaMA3-8B, Qwen1.5-7B, OPT-6.7B and Bloom-7B, as shown in Figure \ref{simq_diff_fig}. Models using rotary position embedding all show \textit{current query is most similar to recent queries}.

\end{document}